\def\year{2022}\relax
\documentclass[letterpaper]{article} 
\usepackage{aaai22}  
\usepackage{times}  
\usepackage{helvet}  
\usepackage{courier}  
\usepackage[hyphens]{url}  
\usepackage{graphicx} 
\usepackage{natbib}  
\usepackage{caption} 
\DeclareCaptionStyle{ruled}{labelfont=normalfont,labelsep=colon,strut=off} 
\usepackage{algorithm}
\usepackage{algorithmic}

\usepackage{newfloat}
\usepackage{listings}
\floatstyle{ruled}
\newfloat{listing}{tb}{lst}{}
\floatname{listing}{Listing}
\copyrighttext{Presented at the AI-HRI Symposium at AAAI Fall Symposium Series (FSS) 2022}
\title{User-specific, Adaptable Safety Controllers Facilitate User Adoption in Human-Robot Collaboration}
\author{
    Ahalya Prabhakar\textsuperscript{\rm 1}, Aude Billard\textsuperscript{\rm 1}}
\affiliations{
    \textsuperscript{\rm 1}École Polytechnique Fédérale de Lausanne \\
    ME A3 495 (Bâtiment ME)\\
    Station 9\\
    CH-1015 Lausanne\\
    \{ahalya.prabhakar, aude.billard\}@epfl.ch
}

\begin{document}

\maketitle

\begin{abstract}
As assistive and collaborative robots become more ubiquitous in the real-world, we need to develop interfaces and controllers that are safe for users to build trust and encourage adoption. In this Blue Sky paper, we discuss the need for co-evolving task and user-specific safety controllers that can accommodate people's safety preferences. We argue that while most adaptive controllers focus on behavioral adaptation, safety adaptation is also a major consideration for building trust in collaborative systems. Furthermore, we highlight the need for adaptation over time, to account for user's changes in preferences as experience and trust builds. We provide a general formulation for what these interfaces should look like and what features are necessary for making them feasible and successful. In this formulation, users provide demonstrations and labelled safety ratings from which a safety value function is learned. These value functions can be updated by updating the safety labels on demonstrations to learn an updated function. We discuss how this can be implemented at a high-level, as well as some promising approaches and techniques for enabling this. 

\end{abstract}

\section{Introduction}
As robots become more capable, collaborative robots are becoming more prevalent in our daily lives. To effectively collaborate with and assist humans, these robots need to be able to adapt to a users' preferences and behavior. However, while industrial robots in factories have been widely adopted across many fields and companies, the adoption of collaborative and assistive robots has not been as successful. This is primarily due to due to safety concerns with humans and robots working in close proximity. While proximity is necessary to improve synergies between human and robots, comfort and safety levels widely vary among skill level and over time. 

In this paper, we focus primarily on users' safety concerns of assistive and collaborative robots. We argue that to successfully address these concerns and encourage adoption by non-expert uses, we need to not only design safe robot controllers according to industry standards, but we also need to design user-specific safe controllers. We define user-specific safe controllers as  ones that take into account the user's beliefs and preferences regarding safe behavior and adapt the robots' resulting behavior accordingly. By doing so, we not only improve robot safety, but we also improve the \emph{users' perception} of robot safety by building trust in the robotic system and encouraging early adoption of these systems by novice users. Furthermore, for enabling long-term adoption, they need to be adaptable to a user's changing needs and preferences over time. For this to be feasible, these systems need to be intuitive to use and efficient to learn. By developing these formulations for enabling human-robot interaction, we can encourage adoption of these systems by a wide variety of users, including novice users who may not be familiar with robotic systems. 

\section{Motivation and Background: Safety Perceptions in Collaborative systems}

Research in technology adoption have long considered safety and trust to be critical factors for enabling technology adoption and operator's ease of use with automation in work places \cite{LEE1994153, grahn2017safety, aaltonen2019experiences}. However, while much of the focus of enabling adoption in collaborative robotics has been focused on safety considerations and industry standards, operator acceptance is also a key component that needs to be addressed \cite{grahn2017safety}. Most research in enabling operator trust in robotic systems identifies reliability and predictability as key factors \cite{desai2009creating, salem2015towards}. In collaborative robotic systems, where users' must inhabit the same physical space and interact with the robotic systems, \emph{perceived safety} is also identified as a key factor, affecting a large number of physiological states and as a result, user comfort \cite{rubagotti2022perceived, bartneck2008measuring, beton2017leader, fosch2019ll, Charalambous2016}. Perceived safety is highlighted as a critical factor for enabling trust and needs to be addressed to enable successful human-robot collaboration in both industrial and assistive robot settings. 
While a positive perception of safety has been identified as a necessary factor for enabling robot acceptance and collaboration, most safety research in collaborative settings only focus on physical safety, rather than perceived safety, and take in user's preferences for task behavior rather than safety standards. Here, we argue that user's preferences should be taken into account in safety design and propose an architecture to enable this. 

\section{Key Features for Safe Human-Robot Interfaces}

\begin{figure*}[h!]
    \centering
    \includegraphics[width=\textwidth]{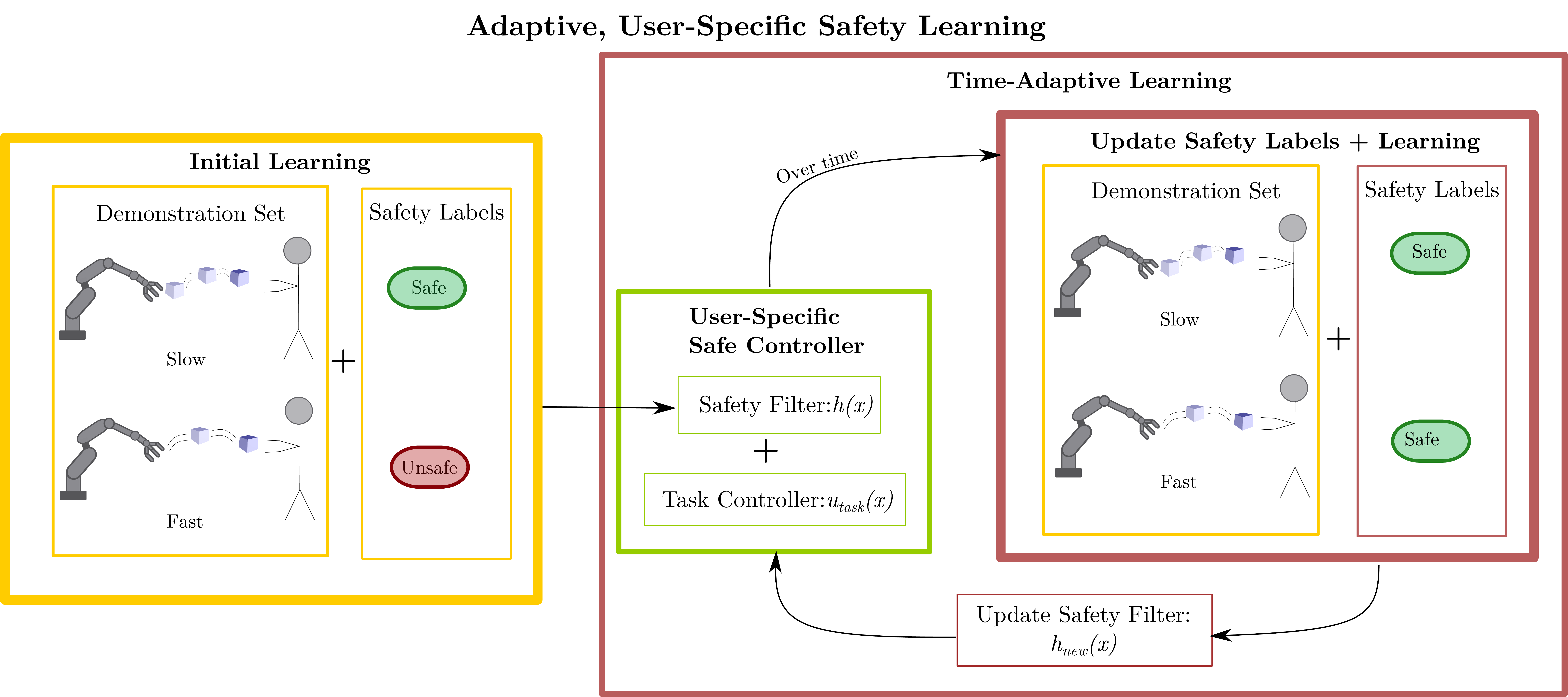}
    \caption{Schematic of Learning Process for Adaptive, User-Specific Safety Controllers with Robot Tossing Example. Users provide labelled datasets of behavior from which a safety filter is learned. Here, they label demonstrations of the robot tossing objects to them with different safety labels. The demonstrations and respective labels are used to learn a safety filter. The learned safe filter is used in conjunction with any task robot controller to ensure successful task performance, while respecting user's safety preferences. Safety preferences are updated over time by updating safety labels to accommodate users' changing preferences as they become more experienced and build trust with the system.}
    \label{fig:schematic}

\end{figure*}

\subsection{User-specific Safety Controllers}
Robots are typically extremely capable and robot safety limits for most tasks may be much higher than allowable around human users---a key reason why most adoption of robots occur in industrial and factory settings where humans are kept at a safe distance from all robots. When we consider safe controllers in the context of human-robot interaction, most work primarily focuses on designing safety in the context of collisions and injury avoidance or task safety. However, in addition to these baseline safety requirements, users may have different preferences with regards to their tolerance for risky behavior that goes beyond baseline human and task safety currently considered in most human-robot interaction settings that may be require additional safety at the expense of task performance. 

A common scenario where these individual preferences are evident is with driving. Some people prefer more cautious behavior--- slower, less aggressive driving--- that can significantly impact task performance, whereas others are more tolerant of task efficient behavior that is faster and more aggressive, which could appear riskier. Not only can seemingly "risky" behavior by a robot cause a more cautious user to not utilize the system, but behavior that is too cautious can cause more risk-tolerant users to be frustrated with the lack of efficiency and also not use it as a result. These safety preferences for behavior not only applies to driving (and self-driving cars), but also to a wide variety of collaborative and assistive robots that are intended to perform a wide variety of tasks. While a large amount of research focuses on behavioral adaption during collaboration, through intent recognition, learning user preferences, and extracting user reward functions, we believe that these adaptive controllers, are necessary but not sufficient, and learning and extracting users' safety preferences specifically are a necessary part of widespread adoption of assistive and collaborative robots. Adapting to these safety preferences can be the determining factor in a users' comfort with the system and their resulting decision to adopt it. 

As such, we propose that user-specific safety filters should be learned for HRI interfaces that can be used in conjunction with task-based robot controllers (that may or may not adapt to users' behaviors). These filters should identify behaviors and states that users' would  consider risky or unsafe and should be used as a baseline safety filter, analogous to standard safe controllers or filters currently used in the control community \cite{ames2019control, xu2015robustness}. These user-specific safety filters could be combined with existing safety filters (i.e., robot- or task-based safety filters) to ensure task safety in conjunction with users' safety tolerances. While these user-specific safety filters may hinder task efficiency or performance, meeting these safety requirements would improve users' trust in these systems. Furthermore, these learned safety filters could be robot-specific or task-specific. In some cases, learning robot-specific safety preferences are sufficient, particularly for robots that are single-use (such as wheelchair robots). However, for robotic systems that are multi-tasking, such as robot arms, assistive robots or prostheses, the safety filters should be task-specific as people's perception of safety with the same robot may differ based on task (e.g., robot pouring water versus cleaning a surface).  

\subsection{Adaptability of Safe Learning}

Importantly, the learning methods should be adaptable over time. Humans learn and become more proficient with tasks and systems as their familiarity increases. Similarly, users may become more risk-tolerant of these systems as their familiarity and trust with them increases. Safety filters should be adaptable over time to accommodate these changing preferences.

To achieve this learning process, we propose a formulation (shown in Figure \ref{fig:schematic}) where users can demonstrate and identify safe, unsafe, and risky behavior according to their preferences with labels accordingly. Learning from Demonstration (LfD) and inverse reinforcement learning (IRL) methods have shown to be intuitive for even novice users, as demonstrations can be an effective way to communicate preferences without requiring the users to have a technical background in robotics. This is particularly relevant for communicating safety preferences, as users may not have the background needed to specify safety limits that may depend on robot kinematics or dynamics. Observing and labelling behavior can be more intuitive than requiring users to provide numerical limits and is a promising approach to communicating this information. Safety filters can be learned from these labelled datasets using inverse reinforcement learning methods, such as \cite{prabhakar2021credit}.

However, these methods can require a large amount of data to accurately learn, which can make them infeasible for real-world deployment. A key focus should be to increase the data efficiency of the learning process, in order to limit the effort on the user to enable learning. There are many research avenues currently being investigated to improve data efficiency for IRL methods such as active learning \cite{sadigh2016information, sadigh2017active, palan2019learning}. One approach for improving learning efficiency could be utilizing active learning methods that specifically query the users' labels for uncertain regions, such as near the boundaries of safe and unsafe behavior where user preferences are most likely to change. Another approach that could be promising for improving efficiency is allowing the robot to demonstrate behaviours (either randomly generated demonstrations or replaying previously demonstrated demonstrations from other users) and only querying users' safety rankings or labels. While this does not improve the data efficiency of the learning algorithm, it does improve the real-world time efficiency for the user, an important distinction. We argue that the time efficiency for learning is a more relevant metric compared to data efficiency for determining feasibility in real-world deployment.  

To enable the formulation to be adaptable to a user's changing preferences, we propose that the user provides updated safety labels to existing demonstrations. By only updating labels, we can significantly reduce the time needed by the user to adapt the safety filter. 
Furthermore, we can actively query for labels close to the boundaries of safety, where the user is most likely to have changed their beliefs about safe and risky behavior. By actively choosing these demonstrations, we can improve the efficiency of the update process, allowing for more frequent and therefore smoother transitions of the safety filter over time. 
By enabling this adaptability over time, we encourage long-term adoption of these collaborative and assistive interfaces.

\section{Conclusion}
Overall, in this paper, we discuss the importance of learning user-specific safety preferences to promote adoption of collaborative and assistive robots by novice users. We explain the need for co-evolving behavior and safety preferences for users over time to promote early adoption by building trust in the safety of the robotic system. In addition, we highlight the need for the formulation to be adaptable over time, to accommodate users' changing preferences as familiarity and proficiency grows to ensure continued use of the systems in the long term. We discuss the importance of these features in assistive and collaborative settings and propose high-level formulation for enabling this. We discuss the main features needed to make this feasible and useful and highlight different approaches that are promising for enabling this.  

\bibliography{references} 
\section{Acknowledgments}
This work was funded by the Swiss National Center of Competence in Research in Robotics (NCCR).
\end{document}